\documentclass[final]{cvpr}

\usepackage{times}
\usepackage{epsfig}
\usepackage{graphicx}
\usepackage{amsmath}
\usepackage{amssymb}

\usepackage{booktabs}
\usepackage{multirow}
\usepackage[vlined,boxed,commentsnumbered,ruled,linesnumbered]{algorithm2e}
\usepackage{amsthm}
\usepackage{comment}

\def\algcomment#1{\textcolor[rgb]{0,0.6,0}{\# #1}}
\newtheorem{theorem}{Theorem}

\newlength{\textfloatsepsave} 
\usepackage[pagebackref=true,breaklinks=true,colorlinks,bookmarks=false]{hyperref}

\def\cvprPaperID{3260} 
\def\confYear{CVPR 2021}

\begin{document}

\title{Combinatorial Learning of Graph Edit Distance via Dynamic Embedding}

\author{
  Runzhong~Wang\textsuperscript{1,2}\qquad Tianqi~Zhang\textsuperscript{1,2}\qquad Tianshu Yu\textsuperscript{3}\qquad Junchi~Yan\textsuperscript{1,2}\thanks{Junchi Yan is the corresponding author.} \qquad Xiaokang~Yang\textsuperscript{2} \\
\textsuperscript{1} Department of Computer Science and Engineering, Shanghai Jiao Tong University \\
\textsuperscript{2} MoE Key Lab of Artificial Intelligence, Shanghai Jiao Tong University\qquad
\textsuperscript{3} Arizona State University\\
{\tt\small \{runzhong.wang,lygztq,yanjunchi,xkyang\}@sjtu.edu.cn \qquad tianshuy@asu.edu }
}

\maketitle

\begin{abstract}
    Graph Edit Distance (GED) is a popular similarity measurement for pairwise graphs and it also refers to the recovery of the edit path from the source graph to the target graph. Traditional A* algorithm suffers scalability issues due to its exhaustive nature, whose search heuristics heavily rely on human prior knowledge. This paper presents a hybrid approach by combing the interpretability of traditional search-based techniques for producing the edit path, as well as the efficiency and adaptivity of deep embedding models to achieve a cost-effective GED solver. Inspired by dynamic programming, node-level embedding is designated in a dynamic reuse fashion and suboptimal branches are encouraged to be pruned. To this end, our method can be readily integrated into A* procedure in a dynamic fashion, as well as significantly reduce the computational burden with a learned heuristic. Experimental results on different graph datasets show that our approach can remarkably ease the search process of A* without sacrificing much accuracy. To our best knowledge, this work is also the first deep learning-based GED method for recovering the edit path.
\end{abstract}

\section{Introduction}
Graph edit distance (GED) is a popular similarity measurement between graphs, which lies in the core of many vision and pattern recognition tasks including image matching~\cite{ChoICCV13}, signature verification~\cite{MaergnerPRL19}, scene-graph edition~\cite{ChenECCV20}, drug discovery~\cite{RiesenSSPR08}, and case-based reasoning~\cite{ZeyenICCBR20}. In general, GED algorithms aim to find an optimal edit path from source graph to target graph with minimum edit cost, which is inherently an NP-complete combinatorial problem~\cite{AbuICPRAM15}:
\begin{equation}
    GED(\mathcal{G}_1, \mathcal{G}_2) = \min_{(e_1, ..., e_l)\in \gamma(\mathcal{G}_1, \mathcal{G}_2)} \sum_{i=1}^l c(e_i)
    \label{eq:ged_form}
\end{equation}
where $\gamma(\mathcal{G}_1, \mathcal{G}_2)$ denote the set of all possible ``edit paths'' transforming source graph $\mathcal{G}_1$ to target graph $\mathcal{G}_2$. $c(e_i)$ measures the cost of edit operation $e_i$.

\begin{figure}[tb!]
    \begin{center}
        \includegraphics[width=\columnwidth]{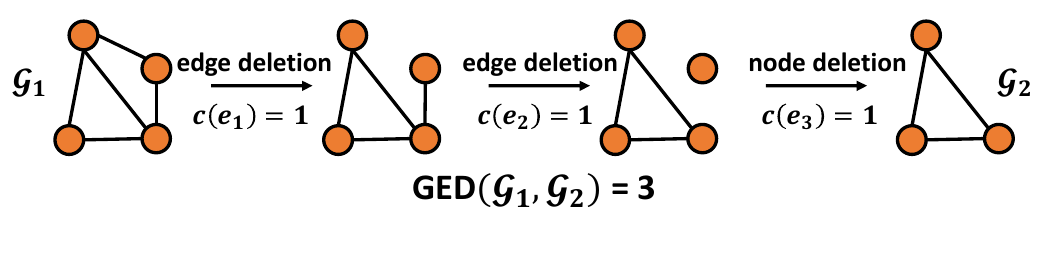}
        \includegraphics[width=\columnwidth]{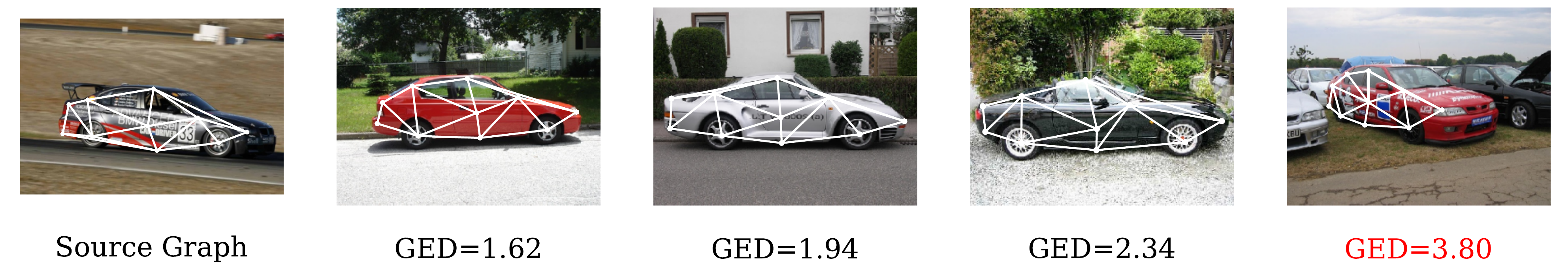}
    \end{center}
    \caption{\textit{Top}: an edit path between two simple graphs $\mathcal{G}_1,\mathcal{G}_2$. \textit{Bottom}: an example of querying images via GED, where only geometric information is involved. The last image shows an ``unsimilar'' image based on GED measurement.}
    \label{fig:edit-graph}
\end{figure}

Exact GED solvers \cite{AbuICPRAM15,RiesenMLG07} guarantee to find the optimal solution under dynamic condition, at the cost of poor scalability on large graphs, and these exact solvers heavily rely on heuristics to estimate the corresponding graph similarity based on the current partial solution. 
Recent efforts in deep graph similarity learning \cite{BaiWSDM19,BaiAAAI20,LiICML19} adopt graph neural networks~\cite{KipfICLR17,ScarselliNN09} to directly regress graph similarity scores, without explicitly incorporating the intrinsic combinatorial nature of GED, hence fail to recover the edit path. However, the edit path is often of the central interest in many applications~\cite{ChenECCV20,ChoICCV13} and most GED works~\cite{AbuICPRAM15,RiesenIVC09,FankGBRPR11,ZengVLDB09,RiesenMLG07} still are more focused on finding the edit path itself.

As the growth of graph size, it calls for more scalable GED solvers which are meanwhile expected to recover the exact edit path. However, these two merits cannot both hold by existing methods. As discussed above, deep learning-based solvers have difficulty in recovering the edit path while the learning-free methods suffer scalability issue. In this paper, we are aimed to design a hybrid solver by combining the best of the two worlds.


Specifically, we resort to A* algorithm~\cite{RiesenMLG07} which is a popular solution among open source GED softwares~\cite{ChangICDE20,RiesenGithub}, and we adopt neural networks to predict similarity scores which are used to guide A* search, in replacement of manually designed heuristics in traditional A*.
We want to highlight our proposed Graph Edit Neural Network (GENN) in two aspects regarding the dynamic programming concepts: Firstly, we propose to reuse the previous embedding information given a graph modification (e.g.\ node deletion) where among the states of A* search tree the graph nodes are deleted progressively\footnote{To distinguish the ``nodes'' in graphs and the ``nodes'' in the search tree, we name ``state'' for the ones in the search tree.}; Secondly, we propose to learn more effective heuristic to avoid unnecessary exploration over suboptimal branches to achieve significant speed-up.

The contributions made in this paper are: 

1) We propose the first (to our best knowledge) deep network solver for GED, where a search tree state selection heuristic is learned by dynamic graph embedding. It outperforms traditional heuristics in efficacy.

2) Specifically, we devise a specific graph embedding method in the spirit of dynamic programming to reuse the previous computation to the utmost extent. In this sense, our method can be naturally integrated with the A* procedure where a dynamical graph similarity prediction is involved after each graph modification, achieving much lower complexity compared to vanilla graph embeddings. 

3) Experimental results on real-world graph data show that our learning-based approach achieves higher accuracy than state-of-the-art manually designed inexact solvers~\cite{FankGBRPR11,RiesenIVC09}. It also runs much faster than A* exact GED solvers~\cite{BergmannInfoSys14,RiesenMLG07} that perform exhaustive search to ensure the global optimum, with comparable accuracy.




\section{Related Work}

\subsection{Traditional GED Solvers}

\noindent\textbf{Exact GED solvers.} For small-scale problems, an exhaustive search can be used to find the global optimum. Exact methods are mostly based on tree-search algorithms such as A* algorithm~\cite{RiesenMLG07}, whereby a priority queue is maintained for all pending states to search, and the visiting order is controlled by the cost of the current partial edit path and a heuristic prediction on the edit distance between the remaining subgraphs~\cite{RiesenIVC09,ZengVLDB09}. 
Other combinatorial optimization techniques, e.g.\ depth-first branch-and-bound~\cite{AbuICPRAM15} and linear programming lower bound~\cite{LerougePR17} can also be adopted to prune unnecessary branches in the searching tree. However, exact GED methods are too time-consuming and they suffer from poor scalability on large graphs~\cite{AbuPR17}.

\noindent\textbf{Inexact GED solvers} aim to mitigate the scalability issue by predicting sub-optimal solutions in (usually) polynomial time. To our knowledge, bipartite matching based methods~\cite{FankGBRPR11,RiesenIVC09,ZengVLDB09} so far show competitive trade-off between time and accuracy, where edge edition costs are encoded into node costs and the resulting bipartite matching problem can be solved in polynomial time by either Hungarian~\cite{KuhnNavalResearch55,RiesenIVC09} or Volgenant-Jonker~\cite{FankGBRPR11,VJ87} algorithm. Beam search~\cite{RiesenGithub} is the greedy version of the exact A* algorithm. Another line of works namely approximate graph matching~\cite{ChoECCV10,JiangNIPS17,WangPAMI17,YanICMR16,YuNIPS18,ZhouCVPR12} are closely related to inexact GED, 
and there are efforts adopting graph matching methods e.g.\ IPFP~\cite{LeordeanuNIPS09} to solve GED problems~\cite{BougleuxICPR16}. Two drawbacks in inexact solvers are that they rely heavily on human knowledge and their solution qualities are relatively poor. 

\subsection{Deep Graph Similarity Learning}
\noindent\textbf{Regression-based Similarity Learning.}
The recent success in machine learning on non-euclidean data (i.e.\ graphs) via GNNs~\cite{FeyCVPR18,KipfICLR17,ScarselliNN09,ZhouArxiv18} has encouraged researchers to design approximators for graph similarity measurements such as GED. SimGNN~\cite{BaiWSDM19} first formulates graph similarity learning as a regression task, where its GCN~\cite{KipfICLR17} and attention~\cite{VaswaniNIPS17} layers are supervised by GED scores solved by A*~\cite{RiesenGithub}. Bai~\textit{et al.}~\cite{BaiAAAI20} extends their previous work by processing a multi-scale node-wise similarity map using CNNs. Li~\textit{et al.}~\cite{LiICML19} propose a cross-graph module in feed-forward GNNs which elaborates similarity learning. Such a scheme is also adopted in information retrieval, where \cite{DaiSIGIR20} adopts a convolutional net to predict the edit cost between texts. However, all these regression models can not predict an edit path, which is mandatory in the GED problem. 

\noindent\textbf{Deep Graph Matching.}
As another combinatorial problem closely related to GED, there is increasing attention in developing deep learning graph matching approaches~\cite{FeyICLR20,JiangArxiv19,WangICCV19} since the seminal work~\cite{ZanfirCVPR18}, and many researchers \cite{RolinekECCV20,WangICCV19,WangArxiv19,YuICLR20} start to take a combinatorial view of graph matching learning rather than a regression task. Compared to graph similarity learning methods, deep graph matching can predict the edit path, but they are designated to match similarly structured graphs and lack particular mechanisms to handle node/edge insertion/deletions. Therefore, modification is needed to fit deep graph matching methods into GED, which is beyond the scope of this paper.

\subsection{Dynamic Graph Embedding}
The major line of graph embedding methods~\cite{FeyCVPR18,KipfICLR17,ScarselliNN09,ZhouArxiv18} assumes that graphs are static which limit their application on real-world graphs that evolve over time.
A line of works namely dynamic graph embedding~\cite{pareja2020evolvegcn,manessi2020dynamic,zheng2019addgraph} aims to solve such issue, whereby recurrent neural networks (RNNs) are typically combined with GNNs to capture the temporal information in graph evolution. 
The applications include graph sequence classification~\cite{manessi2020dynamic}, dynamic link prediction~\cite{pareja2020evolvegcn}, and anomaly detection~\cite{zheng2019addgraph}.
Dynamic graph embedding is also encountered in our GED learning task, however, all these aforementioned works cannot be applied to our setting where the graph structure evolves at different states of the search tree, instead of time steps. 
\begin{figure}[tb!]
    \begin{center}
        \includegraphics[width=0.65\columnwidth]{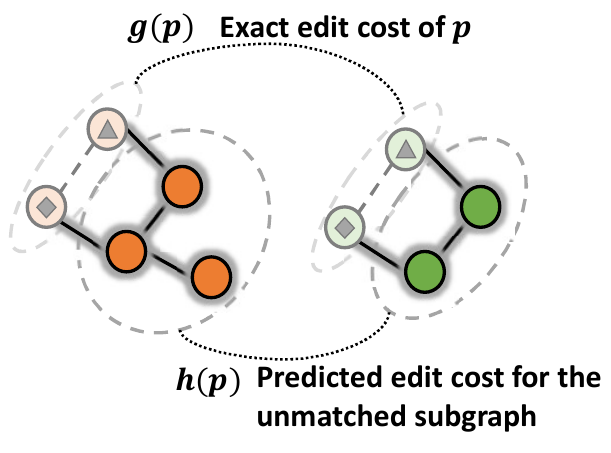}
    \end{center}
    \caption{A partial edit path as one state of A* search tree. Given the partial solution $p=(u_{\blacklozenge}\rightarrow v_{\blacklozenge}, u_{\blacktriangle}\rightarrow v_{\blacktriangle})$, the edge edition $(u_{\blacklozenge}u_{\blacktriangle}\rightarrow v_{\blacklozenge}v_{\blacktriangle})$ can be induced from node editions.}
    \vspace{10pt}
    \label{fig:cost}
\end{figure}
\section{Our Approach}
In this section, we first introduce the A* algorithm for GED in Sec.~\ref{sec:a-star}, then we present our efficient dynamic graph embedding approach GENN for A* in Sec.~\ref{sec:genn}.
\subsection{Preliminaries on A* Algorithm for GED}
\label{sec:a-star}
To exactly solve the GED problem, researchers usually adopt tree-search based algorithms which traverse all possible combinations of edit operations. Among them, A* algorithm is rather popular~\cite{RiesenIVC09,RiesenGithub,RiesenMLG07,ChangICDE20} and we base our learning method on it. In this section, we introduce notations for GED and discuss the key components in A* algorithm. 

GED aims to find the optimal edit path with minimum edit cost, to transform the source graph $\mathcal{G}_1=(V_1, E_1)$ to the target graph $\mathcal{G}_2=(V_2, E_2)$, where $|V_1|=n_1, |V_2|=n_2$. We denote $V_1=\{u_1,...,u_{n_1}\}$, $V_2=\{v_1,...,v_{n_2}\}$ as the nodes in the source graph and the target graph, respectively, and $\epsilon$ as the ``void node''. Possible node edit operations include node substitution $u_i\rightarrow v_j$, node insertion $\epsilon \rightarrow v_j$ and node deletion $u_i \rightarrow \epsilon$, and the cost of each operation is defined by the problem. As shown in Fig.~\ref{fig:cost}, the edge editions can be induced given node editions, therefore only node editions are explicitly considered in A* algorithm.\footnote{Node substitution can be viewed as node-to-node matching between two graphs, and node insertion/deletion can be viewed as matching nodes in source/target graph to the void node, respectively. The concepts ``matching'' and ``edition'' may interchange with each other through this paper.}

\setlength{\textfloatsep}{0pt}
\begin{algorithm}[t!]
\label{alg:a-star}
{\small{
	\caption{\textbf{A* Algorithm for Exact GED}}}}
	\KwIn{Graphs $\mathcal{G}_1=(V_1, E_1)$, $\mathcal{G}_2=(V_2, E_2)$, where $V_1=\{u_1,...,u_{n_1}\}$, $V_2=\{v_1,...,v_{n_2}\}$}
	Initialize \texttt{OPEN} as an empty priority queue;\\
	Insert $(u_1\rightarrow w)$ to \texttt{OPEN} for all $w \in V_2$; \\
	Insert $(u_1 \rightarrow \epsilon)$ to \texttt{OPEN};\\
	\While{no solution is found}
	{
	    Select $p$ with minimum $(g(p)+h(p))$ in \texttt{OPEN}; \\
	    \If{$p$ is a valid edit path}
	    {
	        \textbf{return} $p$ as the solution; \\
	    }
	    \Else
	    {
	        Let $p$ contains $\{u_1, ..., u_k\}\subseteq V_1$ and $W\subseteq V_2$;\\ 
	        \If{$k\leq n_1$}
	        {
	            Insert $p \cup (u_{k+1}\rightarrow v_i)$ to \texttt{OPEN} for all $v_i \in V_2 \backslash W$; \\
	            Insert $p \cup (u_{k+1} \rightarrow \epsilon)$ to \texttt{OPEN}; \\
	        }
	        \Else
	        {
	            Insert $p \cup \bigcup_{v_i\in  V_2 \backslash W}(\epsilon \rightarrow v_i)$ to \texttt{OPEN};\\
	        }
	    }
	}
	\KwOut{An optimal edit path from $\mathcal{G}_1$ to $\mathcal{G}_2$.}
\end{algorithm}

Alg.~\ref{alg:a-star} illustrates a standard A* algorithm in line with \cite{RiesenIVC09,RiesenMLG07}. 
A priority queue is maintained where each state of the search tree contains a partial solution to the GED problem. As shown in Fig.~\ref{fig:cost}, the priority of each state is defined as the summation of two metrics: $g(p)$ representing the cost of the current partial solution which can be computed exactly, and $h(p)$ means the heuristic prediction of GED between the unmatched subgraphs. A* always explores the state with minimum $g(p)+h(p)$ at each iteration and the optimality is guaranteed if $h(p) \leq h^{opt}(p)$ holds for all partial solutions~\cite{RiesenIVC09}, where $h^{opt}(p)$ means the optimal edit cost between the unmatched subgraphs.

\begin{figure*}[tb!]
    \begin{center}
        \includegraphics[width=\textwidth]{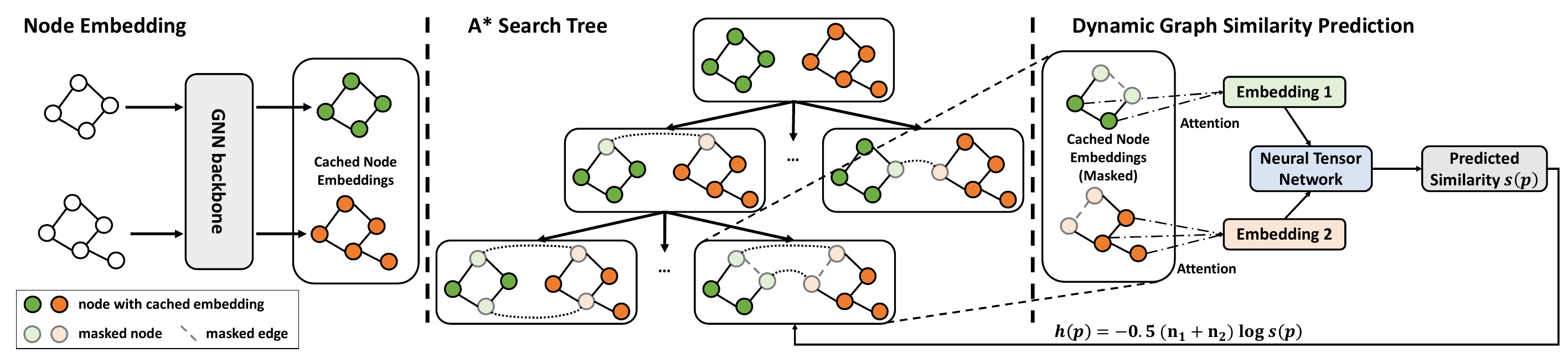}
    \end{center}
    \caption{Our proposed GENN-A*. \textit{Left}: Node embedding. Input graphs are fed into GNN to extract node-level embeddings. These embeddings are cached to be reused in the following computation. \textit{Middle}: A* search tree. The state in the search tree is a matching of nodes between graphs. All matched nodes are masked (light color) and the unmatched subgraphs (dark color) will be involved to predict $h(p)$. \textit{Right}: Dynamic graph similarity prediction. Cached embeddings are loaded for nodes in the unmatched subgraphs, and a graph-level embedding is obtained via attention. Finally the predicted graph similarity $s(p)\in(0, 1)$ is obtained from graph-level embeddings by neural tensor network and transformed to the heuristic score $h(p)$.}
    \label{fig:method}
\end{figure*}

A proper $h(p)$ is rather important to speed up the algorithm, and we discuss three variants of A* accordingly: 
\textbf{1)}~If $h(p)=h^{opt}(p)$, one can directly find the optimal path greedily. However, computing $h^{opt}(p)$ requires another exponential-time solver which is intractable. 
\textbf{2)}~Heuristics can be utilized to predict $h(p)$ where $0\leq h(p) \leq h^{opt}(p)$. Hungarian bipartite heuristic~\cite{RiesenMLG07} is among the best-performing heuristic where the time complexity is $\mathcal{O}((n_1+n_2)^3)$. In our experiments, \textbf{Hungarian-A*}~\cite{RiesenMLG07} is adopted as the baseline traditional exact solver.
\textbf{3)}~{Plain-A*} is the simplest, where it always holds $h(p)=0$ and such strategy introduces no overhead when computing $h(p)$. However, the search tree may become too large without any ``look ahead'' on the future cost.

The recent success of graph similarity learning~\cite{BaiWSDM19,BaiAAAI20,LiICML19} inspires us to predict high-quality $h(p)$ which is close to $h^{opt}(p)$ in a cost-efficient manner via learning. In this paper, we propose to mitigate the scalability issue of A* by predicting $h(p)$ via dynamic graph embedding networks, where $h(p)$ is efficiently learned and predicted and the suboptimal branches in A* are pruned. It is worth noting that we break the optimality condition $h(p)\le h^{opt}(p)$, but the loss of accuracy is acceptable, as shown in experiments.

\subsection{Graph Edit Neural Network}
\label{sec:genn}
An overview of our proposed Graph Edit Neural Network-based A* (GENN-A*) learning algorithm is shown in Fig.~\ref{fig:method}. Our GENN-A* can be split into node embedding module (Sec.~\ref{sec:gnn}), dynamic embedding technique (Sec.~\ref{sec:dynamic}), graph similarity prediction module (Sec.~\ref{sec:similarity}) and finally the training procedure (Sec.~\ref{sec:training}).

\subsubsection{Node Embedding Module}
\label{sec:gnn}
The overall pipeline of our GENN is built in line with SimGNN~\cite{BaiWSDM19}, and we remove the redundant histogram module in SimGNN in consideration of efficiency. Given input graphs, node embeddings are computed via GNNs. 

\textbf{Initialization.} Firstly, the node embeddings are initialized as the one-hot encoding of the node degree. For graphs with node labels (e.g.\ molecule graphs), we encode the node labels by one-hot vector and concatenate it to the degree embedding. The edges can be initialized as weighted or unweighted according to different definitions of graphs.

\textbf{GNN backbone.} Based on different types of graph data, Graph Convolutional Network (GCN)~\cite{KipfICLR17} is utilized for ordinary graph data (e.g.\ molecule graphs and program graphs) and SplineCNN~\cite{FeyCVPR18} is adopted for graphs built from 2D images, considering the recent success of adopting spline kernels to learn geometric features~\cite{FeyICLR20,RolinekECCV20}. The node embeddings obtained by the GNN backbone are cached for further efficient dynamic graph embedding. We build three GNN layers for our GENN in line with \cite{BaiWSDM19}.

\subsubsection{Dynamic Embedding with A* Search Tree}
\label{sec:dynamic}
\begin{figure}[tb!]
    \centering
    \includegraphics[width=\columnwidth]{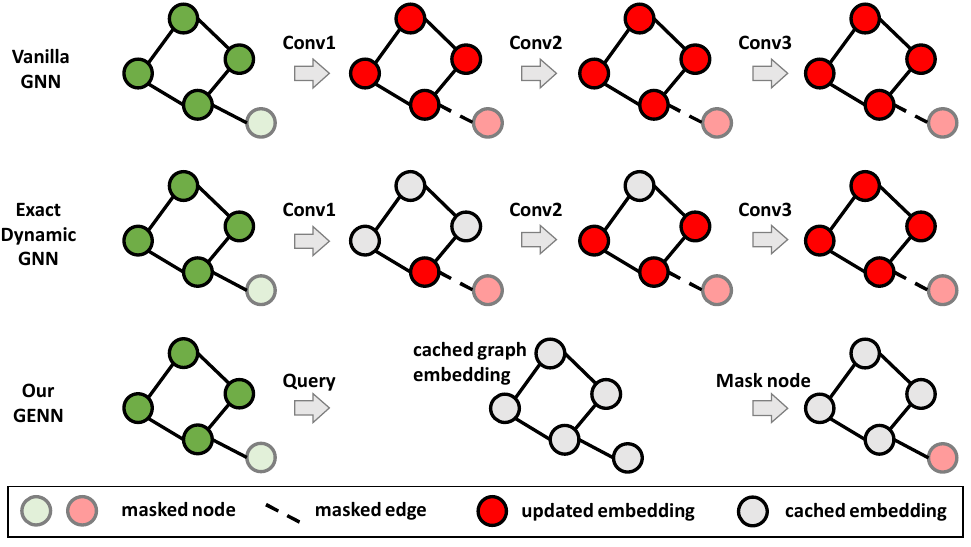}
    \caption{Comparison of three graph neural network variants for dynamic graph embedding in A* algorithm. We assume three graph convolution layers in line with our implementation. In vanilla GNN, a complete forward pass is required for all nodes which contains redundant operations. The exact dynamic GNN caches all intermediate embeddings and only the 3-hop neighbors of the masked node are updated. Finally, our proposed GENN requires no convolution operation and is the most efficient. }
    \vspace{10pt}
    \label{fig:different_gnn}
\end{figure}

A* is inherently a dynamic programming (DP) algorithm where matched nodes in partial solutions are progressively masked. When solving GED, each state of A* contains a partial solution and in our method embedding networks are adopted to predict the edit distance between two unmatched subgraphs. At each state, one more node is masked out in the unmatched subgraph compared to its parent state. Such a DP setting differs from existing so-called dynamic graph embedding problems~\cite{pareja2020evolvegcn,manessi2020dynamic,zheng2019addgraph} and calls for efficient cues since the prediction of $h(p)$ is encountered at every state of the search tree. In this section, we discuss and compare three possible dynamic embedding approaches, among which our proposed GENN is built based on DP concepts.

\textbf{Vanilla GNN.} The trivial way of handling the dynamic condition is that when the graph is modified, a complete feed-forward pass is called for all nodes in the new graph. However, such practice involves redundant computation, which is discussed as follows. We denote $n$ as the number of nodes, $F$ as embedding dimensions, and $K$ as the number of GNN layers. Assuming fully-connected graph as the worst case, the time complexity of vanilla GNN is $\mathcal{O}(n^2FK + nF^2K)$ and no caching is needed.

\textbf{Exact Dynamic GNN.} As shown in the second row of Fig.~\ref{fig:different_gnn}, when a node is masked, only the embeddings of neighboring nodes are affected. If we cache all intermediate embeddings of the forward pass, one can compute the exact embedding at a minimum computational cost. Based on the message-passing nature of GNNs, at the k-th convolution layer, only the k-hop neighbors of the masked node are updated.
However, the worst-case time complexity is still $\mathcal{O}(n^2FK + nF^2K)$ (for fully-connected graphs), and it requires $\mathcal{O}(nFK)$ memory cache for all convolution layers. If all possible subgraphs are cached for best time efficiency, the memory cost grows to $\mathcal{O}(n2^nFK)$ which is unacceptable. Experiment result shows that the speed-up of this strategy is negligible with our testbed.

\textbf{Our GENN.} As shown in the last row of Fig.~\ref{fig:different_gnn}, we firstly perform a forward convolution pass and cache the embeddings of the last convolution layer. During A* algorithm, if some nodes are masked out, we simply delete their embeddings from the last convolution layer and feed the remaining embeddings into the similarity prediction module. Our GENN involves single forward pass which is negligible, and the time complexity of loading caches is simply $\mathcal{O}(1)$ and the memory consumption of caching is $\mathcal{O}(nF)$. 

Our design of the caching scheme of GENN is mainly inspired by DP: given modification on the input graph (node deletion in our A* search case), the DP algorithm reuses the previous results for further computations in consideration of best efficiency. In our GENN, the node embeddings are cached for similarity computation on its subgraphs. In addition, DP algorithms tend to minimize the exploration space for best efficiency, and our learned $h(p)$ prunes sub-optimal branches more aggressively than traditional heuristics which speeds up the A* solver.

\subsubsection{Graph Similarity Prediction}
\label{sec:similarity}
After obtaining the embedding vectors from cache, the attention module and neural tensor network are called to predict the similarity score. For notation simplicity, our discussions here are based on full-sized, original input graphs.


\textbf{Attention module for graph-level embedding.} Given node-level embeddings, the graph-level embedding is obtained through attention mechanism~\cite{VaswaniNIPS17}. We denote $\mathbf{X}_1\in\mathbb{R}^{n_1\times F}, \mathbf{X}_2\in\mathbb{R}^{n_2\times F}$ as the node embeddings from GNN backbone. The global keys are obtained by mean aggregation followed with nonlinear transform:
\begin{equation}
    \bar{\mathbf{X}}_1 = \texttt{mean}(\mathbf{X}_1),
    \bar{\mathbf{X}}_2 = \texttt{mean}(\mathbf{X}_2)
\end{equation}
\begin{equation}
    \mathbf{k}_1 = \texttt{tanh}(\bar{\mathbf{X}}_1 \mathbf{W}_1),
    \mathbf{k}_2 = \texttt{tanh}(\bar{\mathbf{X}}_2 \mathbf{W}_1)
\end{equation}
where $\texttt{mean}(\cdot)$ is performed on the first dimension (node dimension) and $\mathbf{W}_1\in \mathbb{R}^{F\times F}$ is learnable attention weights. Aggregation coefficients are computed from $\mathbf{k}_1, \mathbf{k}_2\in\mathbb{R}^{1\times F}$ and $\mathbf{X}_1, \mathbf{X}_2$:
\begin{equation}
    \mathbf{c}_1 = \delta(\mathbf{X}_1 \mathbf{k}_1^\top \cdot \alpha),
    \mathbf{c}_2 = \delta(\mathbf{X}_2 \mathbf{k}_2^\top \cdot \alpha) 
\end{equation}
where $\alpha=10$ is the scaling factor and $\delta(\cdot)$ means sigmoid. The graph-level embedding is obtained by weighted summation of node embeddings based on aggregation coefficients $\mathbf{c}_1 \in \mathbb{R}^{n_1 \times 1}, \mathbf{c}_2 \in \mathbb{R}^{n_2 \times 1}$:
\begin{equation}
    \mathbf{g}_1 = \mathbf{c}_1^\top \mathbf{X}_1,
    \mathbf{g}_2 = \mathbf{c}_2^\top \mathbf{X}_2 
\end{equation}

\textbf{Neural Tensor Network for similarity prediction.} Neural Tensor Network (NTN)~\cite{SocherNIPS13} is adopted to measure the similarity between $\mathbf{g}_1, \mathbf{g}_2 \in \mathbb{R}^{1\times F}$:
\begin{equation}
    s(\mathcal{G}_1, \mathcal{G}_2) =  f(\mathbf{g}_1\mathbf{W}_2^{[1:t]}\mathbf{g}_2^\top + \mathbf{W}_3 \texttt{cat}(\mathbf{g}_1, \mathbf{g}_2) + \mathbf{b})
\end{equation}
where $\mathbf{W}_2\in \mathbb{R}^{F\times F\times t}, \mathbf{W}_3\in \mathbb{R}^{t \times 2F}, \mathbf{b}\in \mathbb{R}^{t}$ are learnable, the first term means computing $\mathbf{g}_1\mathbf{W}_2[:, :, i]\mathbf{g}_2^\top$ for all $i \in [1...t]$ and then stacking them, $f: \mathbb{R}^t\rightarrow (0,1)$ denotes a fully-connected layer with sigmoid activation, and $\texttt{cat}(\cdot)$ means to concat along the last dimension. $t$ controls the number of channels in NTN and we empirically set $t=16$.

In line with \cite{BaiWSDM19}, the model prediction lies within $(0, 1)$ which represents a normalized graph similarity score with the following connection to GED:
\begin{equation}
    s(\mathcal{G}_1, \mathcal{G}_2) = \exp\left(-{GED(\mathcal{G}_1, \mathcal{G}_2)} \times 2 / {(n_1+n_2)}\right)
    \label{eq:ged_score}
\end{equation}
For partial edit path encountered in A* algorithm, the predicted similarity score $s(p)$ can be transformed to $h(p)$ following Eq.~\ref{eq:ged_score}:
\begin{equation}
    h(p)=- 0.5 (n_1^\prime+n_2^\prime) \log s(p)
\end{equation}
where $n_1^\prime, n_2^\prime$ means the number of nodes in the unmatched subgraph.
The time complexities of attention and NTN are $\mathcal{O}((n_1^\prime+n_2^\prime)F^2)$ and $\mathcal{O}(n_1^\prime n_2^\prime Ft)$, respectively. Since the convolution layers are called only once which is negligible, and the time complexity of loading cached GENN embedding is $\mathcal{O}(1)$, the overall time complexity of each prediction is $\mathcal{O}((n_1^\prime+n_2^\prime)F^2+n_1^\prime n_2^\prime Ft)$. Our time complexity is comparable to the best-known learning-free prediction of $h(p)$~\cite{RiesenMLG07} which is $\mathcal{O}((n_1^\prime+n_2^\prime)^3)$.

\begin{figure*}[tb!]
    \centering
    \includegraphics[width=0.96\textwidth]{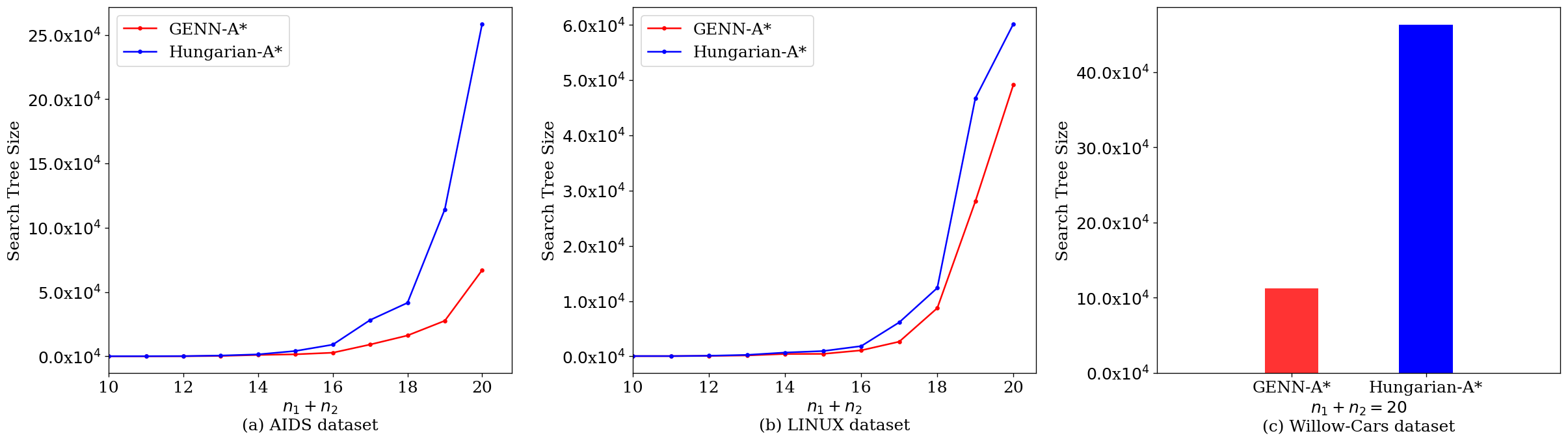}
            \vspace{-3pt}
    \caption{Average search tree size w.r.t. problem size ($n_1+n_2$). The search tree reduces significantly when the problem size grows, especially on more challenging AIDS and Willow-Cars where about $\times 5$ and $\times 4$ reductions of state are achieved respectively via GENN.}
    \label{fig:tree_size}
\end{figure*}

\begin{figure}[tb!]
    \centering
    \includegraphics[width=\columnwidth]{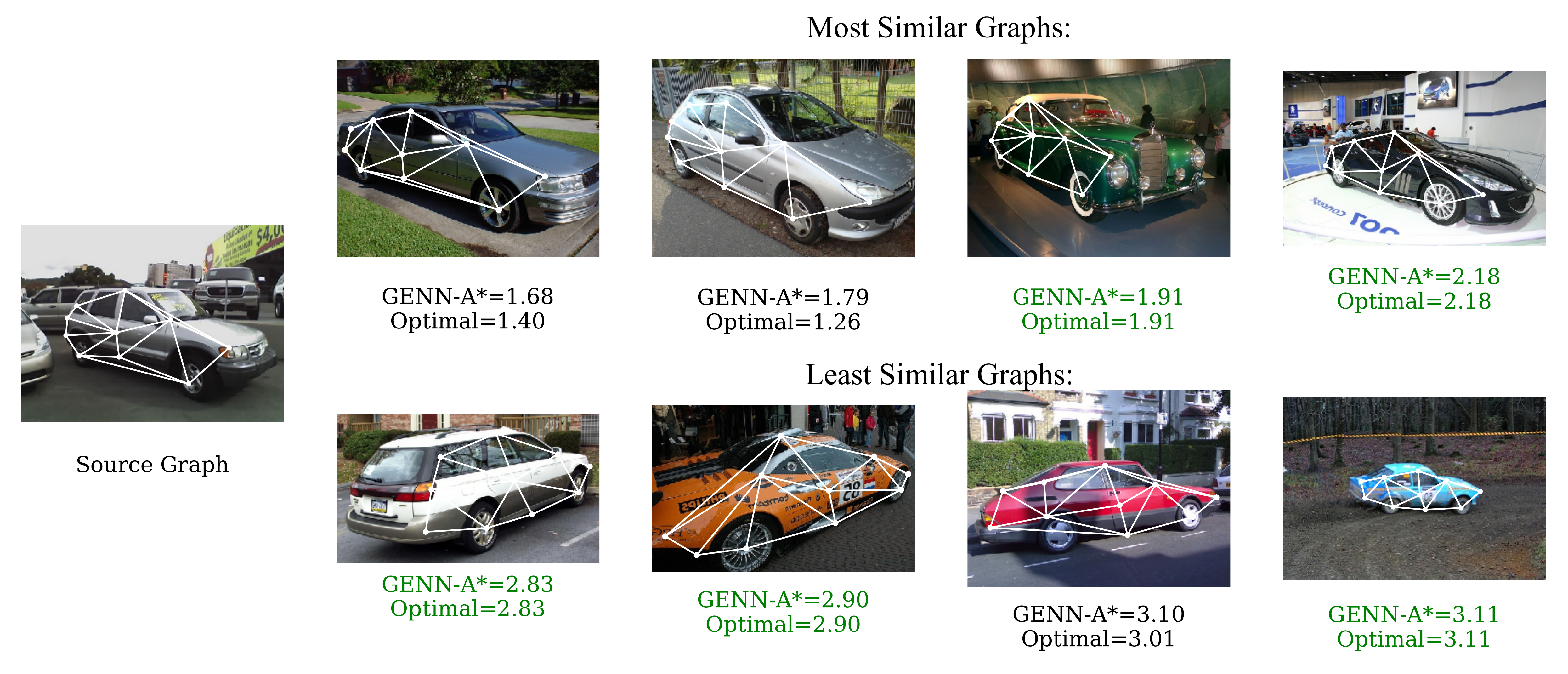}
    \caption{The visualization of a query on Willow-Cars dataset by GENN-A*. All of the 4 most similar graphs are close to the source graph in terms of poses and graph structures, yet the 4 least similar ones vary greatly in their poses and appearances. Green letters mean our GENN-A* solves the optimal GED.}
    \label{fig:willow_visual}
\end{figure}

\begin{algorithm}[tb!]
\label{alg:training}
{\small{
	\caption{\textbf{The Training Procedure of GENN-A*}}}}
	\KwIn{Training set of graphs pairs $\{(\mathcal{G}_i, \mathcal{G}_j)\}$ with similarity score labels $\{s^{gt}(\mathcal{G}_i,\mathcal{G}_j)\}$. }
	\While(\algcomment{training with GT labels}){not converged}
	{
	    Randomly sample $(\mathcal{G}_i,\mathcal{G}_j)$ from training set;\\
	    Compute $s(\mathcal{G}_i,\mathcal{G}_j)$ by vanilla GENN; \\
	    Update parameters by $\texttt{MSE}(s(\mathcal{G}_i,\mathcal{G}_j), s^{gt}(\mathcal{G}_i,\mathcal{G}_j))$;
	}
    \While(\algcomment{finetune with optimal path}){not converged}
	{
        Randomly sample $(\mathcal{G}_i,\mathcal{G}_j)$ from training set;\\
	    Solve the optimal edit path $p^*$ and GED$(p^*)$ by A*; \\
	    Call GENN on $(\mathcal{G}_i,\mathcal{G}_j)$ and cache the embeddings; \\
	    \For{partial edit path $p \subseteq p^*$}
	    {
	        compute $g(p)$ and $h^{opt}(p)=GED(p^*)-g(p)$;\\
	        $s^{opt}(p)=\exp (-2 h^{opt}(p) / (n_1^\prime+n_2^\prime))$;\\
	        compute $s(p)$ from cached GENN embeddings;\\
	        Update parameters by $\texttt{MSE}(s(p), s^{opt}(p))$;
	    }
	}
	\KwOut{GENN with learned parameters.}
\end{algorithm}
\begin{table*}[tb!]
    \centering
    \resizebox{0.96\textwidth}{!}
    {
    \begin{tabular}{r|c|ccc|ccc|ccc}
    \toprule
    \multirow{2}{*}{Method} & Edit & \multicolumn{3}{c|}{AIDS} & \multicolumn{3}{c|}{LINUX} & \multicolumn{3}{c}{Willow-Cars} \\
    \cline{3-11}      &  Path  & mse $(\times10^{-3})$ & $\rho$ & p@10  & mse $(\times10^{-3})$ & $\rho$ & p@10 & mse $(\times10^{-3})$ & $\rho$ & p@10  \\
    \midrule
    SimGNN~\cite{BaiWSDM19} & $\times$ & 1.189  & 0.843  & 0.421  & 1.509  & 0.939  & 0.942 & - & - & - \\
    GMN~\cite{LiICML19}   & $\times$ & 1.886  & 0.751  & 0.401  & 1.027  & 0.933  & 0.833 & - & - & - \\
    GraphSim~\cite{BaiAAAI20} & $\times$ & \textbf{0.787} & 0.874  & 0.534  & \textbf{0.058} & \textbf{0.981}  & \textbf{0.992} & - & - & - \\
    GENN~(ours)  & $\times$ & 1.618  & \textbf{0.901} & \textbf{0.880} & 0.438  & 0.955  & 0.527 & - & - & - \\
    \midrule
    Beam Search~\cite{RiesenGithub} & \checkmark & 12.090 &  0.609 & 0.481 & 9.268 & 0.827  & \textbf{0.973} & 1.820 & 0.815 & 0.725 \\
    Hungarian~\cite{RiesenIVC09} & \checkmark  & 25.296  & 0.510  & 0.360  & 29.805  & 0.638  & 0.913 & 29.936 & 0.553 & 0.650 \\
    VJ~\cite{FankGBRPR11}  & \checkmark   & 29.157  & 0.517  & 0.310  & 63.863  & 0.581  & 0.287  & 45.781 & 0.438 & 0.512 \\
    GENN-A* (ours) & \checkmark   & \textbf{0.635}  &   \textbf{0.959}    &   \textbf{0.871}    & \textbf{0.324}  & \textbf{0.991} & 0.962  & \textbf{0.599} & \textbf{0.928} & \textbf{0.938}\\
    \bottomrule
    \end{tabular}%
    }
    \caption{Evaluation on benchmarks AIDS, LINUX and Willow-Cars. Our method can work either in a way involving explicit edit path generation as traditional GED solvers~\cite{RiesenIVC09,FankGBRPR11,RiesenMLG07}, or based on direct similarity computing without deriving the edit distance~\cite{BaiWSDM19,LiICML19,BaiAAAI20}.
     The evaluation metrics are defined and used by \cite{BaiWSDM19,BaiAAAI20}: \textbf{mse} stands for mean square error between predicted similarity score and ground truth similarity score. $\boldsymbol{\rho}$ means the Spearman's correlation between prediction and ground truth. \textbf{p@10} means the precision of finding the closest graph among the predicted top 10 most similar ones. Willow-Cars is not compared with deep learning methods because optimal GED labels are not available for the training set. The AIDS and LINUX peer method results are quoted from \cite{BaiAAAI20}.}
    \label{tab:aids_and_linux}
\end{table*}

\subsubsection{Supervised Dynamic Graph Learning}
\label{sec:training}
The training of our GENN consists of two steps: Firstly, GENN weights are initialized with graph similarity score labels from the training dataset. Secondly, the model is finetuned with the optimal edit path solved by A* algorithm. The detailed training procedure is listed in Alg.~\ref{alg:training}.

Following deep graph similarity learning peer methods~\cite{BaiWSDM19,BaiAAAI20}, our GENN weights are supervised by ground truth labels provided by the dataset. For datasets with relatively small graphs, optimal GED scores can be solved as ground truth labels. In cases where optimal GEDs are not available, we can build the training set based on other meaningful measurements, e.g. adopting semantic node matching ground truth to compute GED labels.

We further propose a finetuning scheme of GENN to better suit the A* setting. However, tuning GENN with the states of the search tree means we require labels of $h^{opt}(p)$, while solving the $h^{opt}(p)$ for an arbitrary partial edit path is again NP-complete. Instead of solving as many $h^{opt}(p)$ as needed, here we propose an efficient way of obtaining multiple $h^{opt}(p)$ labels by solving the GED only once.

\begin{theorem}
\textbf{(Optimal Partial Cost)} Given an optimal edit path $p^*$ and the corresponding $GED(p^*)$, for any partial edit path $p \subseteq p^*$, there holds $g(p)+h^{opt}(p)=GED(p^*)$.
\label{theorem:hp}
\end{theorem}
\begin{proof}
If $g(p) + h^{opt}(p) > GED(p^*)$, then the minimum edit cost following $p$ is larger than $GED(p^*)$, therefore $p$ is not a partial optimal edit path, which violates $p \subseteq p^*$. 
If $g(p) + h^{opt}(p) < GED(p^*)$, it means that there exists a better edit path whose cost is smaller than $GED(p^*)$, which violates the condition that $p^*$ is the optimal edit path.
Thus, $g(p) + h^{opt}(p) = GED(p^*)$.
\end{proof} 
Based on Theorem~\ref{theorem:hp}, there holds $h^{opt}(p)=GED(p^*)-g(p)$ for any partial optimal edit path. Therefore, if we solve an optimal $p^*$ with $m$ node editions, $(2^m-1)$ optimal partial edit paths can be used for finetuning. In experiments, we randomly select 200 graph pairs for finetuning since we find it adequate for convergence.

\section{Experiment}

\subsection{Settings and Datasets}
\label{sec:exp_setting}
We evaluate our learning-based A* method on three challenging real-world datasets: AIDS, LINUX~\cite{WangICDE12}, and Willow dataset~\cite{ChoICCV13}. 

\noindent\textbf{AIDS dataset} contains chemical compounds evaluated for the evidence of anti-HIV activity\footnote{\url{https://wiki.nci.nih.gov/display/NCIDTPdata/AIDS+Antiviral+Screen+Data}}. AIDS dataset is pre-processed by \cite{BaiWSDM19} who remove graphs more than 10 nodes and the optimal GED between any two graphs is provided. Following \cite{BaiWSDM19}, we define the node edition cost $c(u_i\rightarrow v_j)=1$ if $u_i, v_j$ are different atoms, else $c(u_i\rightarrow v_j)=0$. The node insertion and deletion costs are both defined as 1. The edges are regraded as non-attributed, therefore edge substitution cost $=0$ and edge insertion/deletion cost $= 1$. 

\noindent\textbf{LINUX dataset} is proposed by \cite{WangICDE12} which contains Program Dependency Graphs (PDG) from the LINUX kernel, and the authors of \cite{BaiWSDM19} also provides a pre-processed version where graphs are with maximum 10 nodes and optimal GED values are provided as ground truth. All nodes and edges are unattributed therefore the substitution cost is 0, and the insertion/deletion cost is 1. 

\noindent\textbf{Willow dataset} is originally proposed by \cite{ChoICCV13} for semantic image keypoint matching problem, and we validate the performance of our GENN-A* on computer vision problems with the Willow dataset. All images from the same category share 10 common semantic keypoints. ``Cars'' dataset is selected in our experiment. With Willow-Cars dataset, graphs are built with 2D keypoint positions by Delaunay triangulation, and the edge edition cost is defined as $c(\mathcal{E}_i\rightarrow\mathcal{E}_j)=|\mathcal{E}_i - \mathcal{E}_j|$ where $\mathcal{E}_i, \mathcal{E}_j$ are the length of two edges. Edge insertion/deletion costs of $\mathcal{E}_i$ are defined as $|\mathcal{E}_i|$. All edge lengths are normalized by 300 for numerical concerns. The node substitution has 0 cost, and $c(u_i\rightarrow \epsilon)=c(\epsilon\rightarrow v_j)=\infty$ therefore node insertion/deletion are prohibited. We build the training set labels by computing the GED based on semantic keypoint matching relationship, and it is worth noting such GEDs are different from the optimal ones. However, experiment results show that such supervision is adequate to initialize the model weights of GENN. 

Among all three datasets, LINUX has the simplest definition of edit costs. In comparison, AIDS has attributed nodes and Willow dataset has attributed edges, making these two datasets more challenging than LINUX dataset. In line with \cite{BaiWSDM19}, we split all datasets by 60\% for training, 20\% for validation, and 20\% for testing.

Our GENN-A* is implemented with Pytorch-Geometric~\cite{FeyICLR19} and the A* algorithm is implemented with Cython~\cite{Cython} in consideration of performance. We adopt GCN~\cite{KipfICLR17} for AIDS and LINUX datasets and SplineCNN~\cite{FeyCVPR18} for 2D Euclidean data from Willow-Cars (\#kernels=16). The number of feature channels are defined as 64, 32, 16 for three GNN layers. Adam optimizer~\cite{adam} is used with 0.001 learning rate and $5\times 10^{-5}$ weight decay. We set batch size=128 for LINUX and AIDS, and 16 for Willow. All experiments are run on our workstation with Intel i7-7820X@3.60GHz and 64GB memory. Parallelization techniques e.g.\ multi-threading and GPU parallelism are not considered in our experiment. 

\begin{table}[tb!]
    \centering
    \resizebox{0.9\columnwidth}{!}
    {
    \begin{tabular}{r|ccc}
    \toprule
    Method & AIDS  & LINUX & Willow-Cars \\
    \midrule
    Hungarian-A*~\cite{RiesenMLG07} & 29.915 & 2.332 & 188.234 \\
    GENN-A*~(ours) & \textbf{13.323} & \textbf{2.177} & \textbf{78.481} \\
    \bottomrule
    \end{tabular}%
    }
    \caption{Averaged time (sec) for solving GED problems.}
    \label{tab:time}
\end{table}

\begin{table}[tb!]
    \centering
    \resizebox{\columnwidth}{!}{
    \begin{tabular}{r|ccc|c}
    \toprule
          & {Vanilla GNN} & {Exact Dynamic GNN} & {GENN (ours)} & {Hungarian~\cite{RiesenMLG07}} \\
    \midrule
    time & 2.329 & 3.145 & 0.417 & 0.358 \\
    \bottomrule
    \end{tabular}%
    }
    \caption{Averaged time (msec) of different methods to predict $h(p)$. Statistics are collected on LINUX dataset.}
    \vspace{10pt}
    \label{tab:gnn_time}
\end{table}

\subsection{Peer Methods}
\textbf{Hungarian-A*}~\cite{RiesenMLG07} is selected as the exact solver baseline, where Hungarian bipartite matching is used to predict $h(p)$. We reimplement Hungarian-A* based on our Cython implementation for fair comparison. We also select \textbf{Hungarian solver}~\cite{RiesenIVC09} as the traditional inexact solver baseline in our experiments. It is worth noting that Hungarian bipartite matching can be either adopted as heuristic in A* algorithm (Hungarian heuristic for A*), or to provide a fast sub-optimal solution to GED (Hungarian solver), and readers should distinguish between these two methods. Other inexact solvers are also considered including \textbf{Beam search}~\cite{RiesenGithub} which is the greedy version of A* and \textbf{VJ}~\cite{FankGBRPR11} which is an variant from Hungarian solver. 

\begin{figure*}[tb!]
    \centering
    \includegraphics[width=\textwidth]{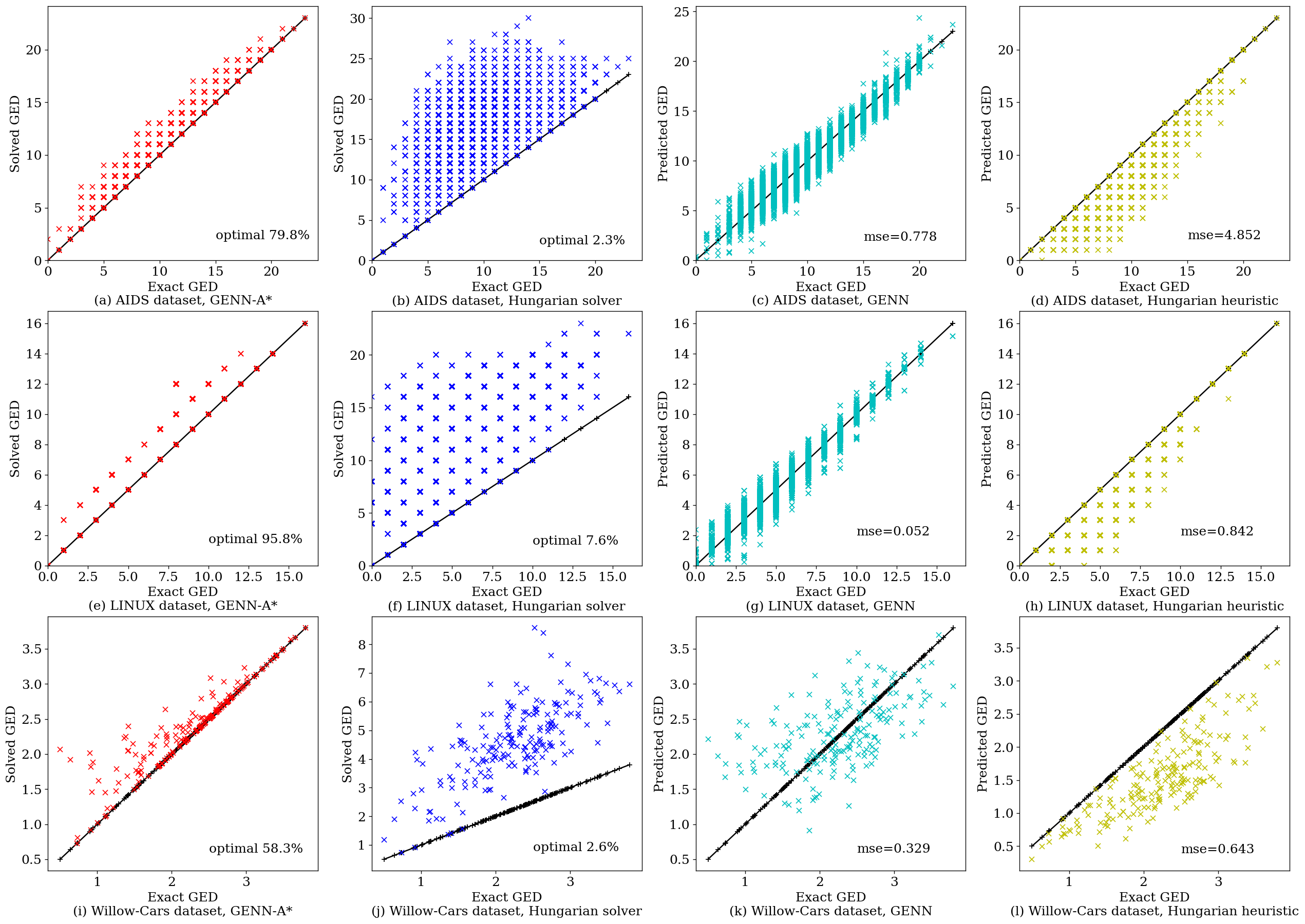}
    \caption{The scatter plots of our proposed GENN-A* (red), inexact Hungarian solver~\cite{RiesenIVC09} (blue, upper bound), our GENN network (cyan) and Hungarian heuristic for A*~\cite{RiesenMLG07} (yellow, lower bound) on AIDS, LINUX and Willow-Cars datasets. The left two columns are GED solvers and the right two columns are methods used to predict $h(p)$ in A* algorithm. Every dot is plotted with optimal GED value on x-axis and the solved (or predicted) GED value on y-axis. Optimal black dots are plotted as references. Our GENN-A* (red) achieves tighter upper bounds than inexact Hungarian solver~\cite{RiesenIVC09} (blue), where a significant amount of problems are solved to optimal. Our regression model GENN (cyan) also predicts more accurate $h(p)$ than Hungarian heuristic~\cite{RiesenMLG07} (yellow), resulting in reduced search tree size of GENN-A* compared to Hungarian-A*.
    }
    \label{fig:scatter_plot}
\end{figure*}

For regression-based deep graph similarity learning methods, we compare \textbf{SimGNN}~\cite{BaiWSDM19}, \textbf{GMN}~\cite{LiICML19} and \textbf{GraphSim}~\cite{BaiAAAI20}. Our GENN backbone can be viewed as a simplified version from these methods, because the time efficiency with dynamic graphs is our main concern.

\subsection{Results and Discussions}

The evaluation of AIDS, LINUX, and Willow-Cars dataset in line with \cite{BaiAAAI20} is presented in Tab.~\ref{tab:aids_and_linux}, where the problem is defined as querying a graph in the test dataset from all graphs in the training set. The similarity score is defined as Eq.~\ref{eq:ged_score}. Our regression model GENN has comparable performance against state-of-the-art with a simplified pipeline, and our GENN-A* best performs among all inexact GED solvers. We would like to point out that \textbf{mse} may not be a fair measurement when comparing GED solvers with regression-based models: Firstly, GED solvers can predict edit paths while such a feature is not supported by regression-based models. Secondly, the solutions of GED solvers are upper bounds of the optimal values, but regression-based graph similarity models~\cite{BaiWSDM19,BaiAAAI20,LiICML19} predicts GED values on both sides of the optimums. Actually, one can reduce the \textbf{mse} of GED solvers by adding a bias to the predicted GED values, which is exactly what the regression models are doing. 

The number of states which have been added to $\texttt{OPEN}$ in Alg.~\ref{alg:a-star} is plotted in Fig.~\ref{fig:tree_size}, where our GENN-A* significantly reduces the search tree size compared to Hungarian-A*. Such search-tree reduction results in the speed-up of A* algorithm, as shown in Tab.~\ref{tab:time}. Both evidences show that our GENN learns stronger $h(p)$ than Hungarian heuristic~\cite{RiesenMLG07} whereby redundant explorations on suboptimal solutions are pruned. We further compare the inference time of three discussed dynamic graph embedding method in Tab.~\ref{tab:gnn_time}, where our GENN runs comparatively fast against Hungarian heuristic, despite the overhead of calling PyTorch functions from Cython. Exact Dynamic GNN is even slower than the vanilla version, since its frequent caching and loading operations may consume additional time. 
It is worth noting that further speedup can be achieved by implementing all algorithms in C++ and adopting parallelism techniques, but these may be beyond the scope of this paper. 

In Fig.~\ref{fig:scatter_plot} we show the scatter plot of GENN-A* and inexact Hungarian solver~\cite{RiesenIVC09} as GED solvers, as well as GENN and Hungarian heuristic as the prediction methods on $h(p)$. Our GENN-A* benefits from the more accurate prediction of $h(p)$ by GENN, solving the majority of problem instances to optimal. We also visualize a query example on Willow-Car images in Fig.~\ref{fig:willow_visual} done by our GENN-A*.


\section{Conclusion}
This paper has presented a hybrid approach for solving the classic graph edit distance (GED) problem by integrating a dynamic graph embedding network for similarity score prediction into the edit path search procedure. Our approach inherits the good interpretability of classic GED solvers as it can recover the explicit edit path between two graphs while it achieves better cost-efficiency by replacing the manual heuristics with the fast embedding module. Our learning-based A* algorithm can reduce the search tree size and save running time, at the cost of little accuracy lost. 

\section*{Acknowledgments}
This research was supported by China Major State Research Development Program (2020AAA0107600), NSFC (61972250, U19B2035).

{\small
\bibliographystyle{ieee_fullname}
\bibliography{egbib}
}

\end{document}